# ROBUST UNSUPERVISED FLEXIBLE AUTO-WEIGHTED LOCAL-COORDINATE CONCEPT FACTORIZATION FOR IMAGE CLUSTERING

*Zhao Zhang [1,4], Yan Zhang [1], Sheng Li [2], Guangcan Liu [3], Meng Wang [4], Shuicheng Yan [5]*

[1] Soochow University, [4] Hefei University of Technology, [2] Adobe Research, [3] Nanjing University of Information Science and Technology, [5] National University of Singapore

**ABSTRACT**

We investigate the high-dimensional data clustering problem by proposing a novel and unsupervised representation learning model called Robust Flexible Auto-weighted Local-coordinate Concept Factorization (RFA-LCF). RFA-LCF integrates the robust flexible CF, robust sparse local-coordinate coding and the adaptive reconstruction weighting learning into a unified model. The adaptive weighting is driven by including the joint manifold preserving constraints on the recovered clean data, basis concepts and new representation. Specifically, our RFA-LCF uses a L2,1-norm based flexible residue to encode the mismatch between clean data and its reconstruction, and also applies the robust adaptive sparse local-coordinate coding to represent the data using a few nearby basis concepts, which can make the factorization more accurate and robust to noise. The robust flexible factorization is also performed in the recovered clean data space for enhancing representations. RFA-LCF also considers preserving the local manifold structures of clean data space, basis concept space and the new coordinate space jointly in an adaptive manner way. Extensive comparisons show that RFA-LCF can deliver enhanced clustering results.

## 1. INTRODUCTION

In recent decades, lots of effective matrix factorization based models have been proposed for data presentation [1][34-35], of which Singular Value Decomposition (SVD) [5], Nonnegative Matrix Factorization (NMF) [8], Vector quantization (VQ) [6], and Concept Factorization (CF) [24] are several representative methods [32]. Among these factorization methods, NMF and CF differ from the others since they imposes the nonnegative constraints on the factorization matrices explicitly.

Due to the additive nonnegative constraints, both NMF and its variants, for instance Projective NMF (PNMF) [28], Graph Regularized NMF (GNMF) [2], Constrained NMF (CNMF) [10], Dual-graph Sparse NMF (DSNMF) [13], Graph dual regularization NMF (DNMF) [18], and Parameter-less Auto-weighted Multiple GNMF (PAMGNMF) [19] are widely used for learning parts-based representation for representing and clustering data. But NMF and variants cannot deal with data in kernel space directly. To handle this issue, CF and its variants were proposed, for instance Locally Consistent CF (LCCF) [3], Local Coordinate CF (LCF) [11], Graph-regularized CF with Local Coordinate (LGCF) [9], Dual-graph regularized CF (GCF) [26] and Graph-Regularized LCF (GRLCF) [27].

Although the enhanced representation results are produced by aforementioned manifold preserving CF variants, they still suffer from several obvious drawbacks. First, to preserve the manifold locality of representations, LCCF, GRLCF, LGCF and GCF usually search the neighbors of each sample using $k$-neighborhood or $\varepsilon$-neighborhood firstly, and then pre-compute the graph weights by a separable step before factorization. But estimating an optimal $k$ or $\varepsilon$ still remains a tricky issue in real applications [17][22], and using the same $k$ or $\varepsilon$ value for each sample is also unreasonable since real application data usually have complex distributions [1]. Moreover, the pre-calculated weights also cannot be ensured to be optimal for calculating the new representation of original data explicitly. Second, real data usually has noise, redundant information and unfavorable features that may cause negative effects on the results. Thus, it would be better to weight and represent data in the recovered clean space, which can potentially produce more accurate and compact data representations. Third, although CF and variants have used a residue term to minimize the reconstruction error between the original data and the product of three factors as a hard constraint for discovering the new representation, we still argue that such operation assumes that a linear combination of the cluster centers should be able to represent each data point, but these hard constraints may be over-strict in the practical applications. For example, some real-world application dataset may have the nonlinear manifold structures. In such case, the results by the linear reconstruction may be inaccurate.

In this paper, we therefore propose a novel unsupervised representation method termed Robust Flexible Auto-weighted Local-coordinate Concept Factorization (RFA-LCF). The idea of RFA-LCF is to enhance the data representation ability by explicitly improving the robust properties of factorization to noise and error by jointly recovering the underlying clean data, enhancing the similarity by adaptive weighting, and providing flexible residue for encoding the mismatch between data and the product by relaxing the hard constraint on the residue.

For the robust flexible factorization, RFA-LCF improves the data representation in twofold. First, it enhances the robust properties against noise and gross errors by involving an error correction process, and then conducts the factorization in the recovered clean data space for enhancing the representations. Besides, RFA-LCF uses the sparse L2,1-norm to encode the reconstruction loss between the recovered clean data and its reconstruction, since L2,1-norm is robust to noise and outliers and moreover has potential to minimize the loss [26]. Second, RFA-LCF considers relaxing the aforementioned mismatch to handle the data sampled from a nonlinear manifold, inspired by [16]. That is, RFA-LCF applies a flexible penalty term on the factorization loss by relaxing the existing assumption that each data point might be represented by a linear combination of the cluster centers, which is clearly a soft constraint.

To guarantee the encoded locality and sparsity to be more accurate, RFA-LCF also integrates the adaptive reconstruction weighting with the robust flexible CF to discover the manifold structures of given data, basis concepts and new representation in an adaptive manner at the same time. Moreover, RFA-LCF also uses the robust adaptive sparse local coordinate coding performed in clean data space to represent the data by using a few most nearby basis concepts, which is different from LCF that performs the local coordinate coding in the original data space. Thus, the new representation by our RFA-LCF will be potentially more accurate, informative and robust to noise.

## 2. PROBLEM FORMULATION

We present the formulation of RFA-LCF. Given the dataset $X = [x_1, x_2, ..., x_N] \in \mathbb{R}^{d \times N}$, RFA-LCF jointly obtains a L2,1-norm based sparse projection $P \in \mathbb{R}^{d \times d}$ to remove noise and outliers in the data by embedding $X$ onto it directly, and then runs the factorization over the recovered clean data $P^T X$ to calculate two nonnegative matrices $W \in \mathbb{R}^{N \times R}$ and $V^T \in \mathbb{R}^{R \times N}$ so that the product of $X$, $W$ and $V^T$, i.e., $XWV^T \in \mathbb{R}^{d \times N}$, can approximate the recovered clean data $P^T X$. Clearly, RFA-LCF performs concept factorization in the clean data space spanned by using $P$ rather than the original input space $X$, which can potentially make the factorization process more accurate and robust. For robust flexible CF, RFA-LCF set the factorization based on the clean data as $P^T X = \hbar(X, W, V) + P_0^T X$, where $\hbar(X, W, V)$ is a transform function for factorizing data. Assuming that $\hbar(X, W, V)$ is the linear regression function $X^T P + eb^T$, where $e = (1,1,...,1)^T$ is a column vector of all ones and $b \in \mathbb{R}^{d \times 1}$ is the bias vector, then $P_0^T X$ can encode the mismatch between $X^T P + eb^T$ and $XWV^T$. To encode the reconstruction residue $P_0^T X$ more accurate, sparse L2,1-norm is imposed on it, i.e., $\|X^T P + eb^T - VW^T X^T\|_{2,1}$. To encode neighborhood information and pairwise similarities more accurately, RFA-LCF encodes the manifold structures jointly over the clean data $P^T X$, basis concept vectors $XW$ and new coordinates $V^T$ in an adaptive manner by minimizing the joint reconstruction error them explicitly, that is, $\|P^T X - P^T XQ\|_F^2 + \|W^T - W^T Q\|_F^2 + \|V^T - V^T Q\|_F^2$, where $Q$ is the reconstruction weight matrix. RFA-LCF also involves the robust adaptive neighborhood preserving local coordinate coding to represent data using a few most nearby basis concepts, which can potentially make the factorization result more informative. These discussions can lead to the following objective function for RFA-LCF:

$$O = \|X^T P + eb^T - VW^T X^T\|_{2,1} + \alpha f(W,V) + \beta g(Q) + \gamma \|P\|_{2,1}, \quad (1)$$
$$s.t. \ W, V, Q \geq 0, \ Q_{ii} = 0$$

where $W, V, Q \geq 0$ are nonnegative constraints, $Q_{ii} = 0$ is added to avoid the trivial solution $Q = I$, and $\alpha, \beta, \gamma > 0$ are trade-off parameters. Since L2,1-norm forces the residue to be sparse in rows and robust to noise [25], so minimizing the L2,1-norm based residue has a potential to reduce the reconstruction error. $f(W,V)$ is the robust adaptive locality and sparsity constraint term and $g(Q)$ is auto-weighted learning term, defined as

$$f(W,V) = \sum_{i=1}^{N} \sum_{r=1}^{R} |v_{ri}^T| \left\| \sum_{j=1}^{N} w_{jr} P^T x_j - P^T x_i \right\|^2$$

$$g(Q) = \left\| \begin{pmatrix} P^T X \\ W^T \\ V^T \end{pmatrix} - \begin{pmatrix} P^T X \\ W^T \\ V^T \end{pmatrix} Q \right\|_F^2 . \quad (2)$$

To highlight the benefits of involving $f(W,V)$ and $g(Q)$, next we briefly discuss the sum of them as follows:

$$\alpha f(W,V) + \beta g(Q) = \alpha \sum_{i=1}^{N} \sum_{r=1}^{R} |v_{ri}^T| \left\| \sum_{j=1}^{N} w_{jr} P^T x_j - P^T x_i \right\|^2$$

$$+ \left\| \begin{pmatrix} \sqrt{\beta} P^T X \\ \sqrt{\beta} W^T \\ \sqrt{\beta} V^T \end{pmatrix} - \begin{pmatrix} \sqrt{\beta} P^T X \\ \sqrt{\beta} W^T \\ \sqrt{\beta} V^T \end{pmatrix} Q \right\|_F^2 , \quad (3)$$

from which one can find that the neighborhood relationship can also be encoded in an adaptive manner by integrating the reconstruction error $\|\sqrt{\beta} W^T - \sqrt{\beta} W^T Q\|_F^2 + \|\sqrt{\beta} V^T - \sqrt{\beta} V^T Q\|_F^2$ based on the basis concept vectors $W^T$ and coordinates $V^T$ into local coordinate coding. Besides, RFA-LCF performs encodes the coordinates in the clean data space, thus RFA-LCF involves a robust adaptive neighborhood preserving locality and sparsity constraint penalty between the anchor point $u_r$ and $x_i$.

## 3. OPTIMIZATION

RFA-LCF has several variables in RFA-LCF and they depend on each other, so we follow the common procedures to update them alternately. Let $Y = X^T P + eb^T - VW^T X^T = [y^1, y^2, ..., y^N]$, $O$ be the objective function, and $M \in \mathbb{R}^{N \times N}$ be a diagonal matrix with entries $m_{ii} = 1/(2\|y^i\|_2), i = 1, 2, ..., N$. Based on the property of L2,1-norm [25], we can have

$$\|X^T P + eb^T - VW^T X^T\|_{2,1}$$
$$= 2tr\left((P^T X + be^T - XWV^T)M(X^T P + eb^T - VW^T X^T)\right). \quad (4)$$

By taking the derivative of Eq.(3) w.r.t. bias $b$ and setting the derivative to zero, we can easily obtain

$$b = (XWV^T Me - P^T XMe)/N_M^+, \quad (5)$$

where $N_M^+ = e^T Me = \sum_i m_{ii}$ is a constant. So, the flexible residue $\wp_{loss}$ can be rewritten as

$$\wp_{loss} = X^T P + eb^T - VW^T X^T$$
$$= X^T P + e(e^T MVW^T X^T - e^T MX^T P)/N_M^+ - VW^T X^T$$
$$= H_e X^T P - (VW^T X^T - ee_M^T VW^T X^T / N_M^+) , \quad (6)$$
$$= H_e X^T P - H_e VW^T X^T$$

where $H_e = I - ee_M^T / N_M^+$ and $e_M^T = e^T M = [m_{11}, m_{22}, ..., m_{NN}] \in \mathbb{R}^N$. Then we have $\|P\|_{2,1} = tr(2P^T SP)$ and $\|H_e X^T P - H_e VW^T X^T\|_{2,1} = $

$tr(2(P^TX - XWV^T)H_e^T MH_e(X^TP - VW^TX^T))$, where $S$ is a diagonal matrix with entries $S_{ii} = 0.5/\|P^i\|_2$, $P^i$ is the $i$-th row vector of $P$. Suppose that each $(H_e X^T P - H_e VW^T X^T)^i \neq 0$ and $P^i \neq 0$ over each $i$, and let $H = (X^T P, W, V)^T$, we have the following matrix trace based formulation for RFA-LCF:

$$\min_{W,V,Q,P,M,S} tr((P^TX - XWV^T)H_e^T MH_e(X^TP - VW^TX^T))$$
$$+ \alpha tr\left(\sum_{i=1}^N (P^T x_i e^T - P^T XW) L_i (P^T x_i e^T - P^T XW)^T\right), \quad (7)$$
$$+ \beta tr(HGH^T) + \gamma tr(P^T SP)$$
$$s.t.\ W, V, Q \geq 0,\ Q_{ii} = 0$$

where $L_i = diag(|v_i|)$ and $G = (I-Q)(I-Q)^T$. As a result, the optimization of our RFA-LCF can be described as follows:

**1) Fix *others*, update the matrix factors *W, V*.**
Let $\psi_{jr}$ and $\phi_{jr}$ be Lagrange multipliers for constraints $W \geq 0$, and $V \geq 0$ respectively, $\Psi = [\psi_{jr}]$ and $\Phi = [\phi_{jr}]$, then Lagrange function $\mathcal{L}_1$ of Eq.(7) can be constructed as

$$\mathcal{L}_1 = tr((P^TX - XWV^T)H_e^T MH_e(X^TP - VW^TX^T))$$
$$+ \alpha tr\left(\sum_{i=1}^N (P^T x_i e^T - P^T XW) L_i (P^T x_i e^T - P^T XW)^T\right). \quad (8)$$
$$+ \beta tr(HGH^T) + tr(\Psi W^T) + tr(\Phi V)$$

By taking the derivatives of $\mathcal{L}_1$ w.r.t $W$ and $V$, and using the Karush-Kuhn-Tucker conditions [33] $\psi_{jr} w_{jr} = 0$, and $\phi_{jr} v_{jr} = 0$, we can easily obtain the following updating rules:

$$w_{jr} \leftarrow w_{jr} \frac{\left(X^T P^T XCV + \alpha \sum_{i=1}^N X^T PP^T x_i e^T L_i\right)_{jr}}{\left(KWV^T CV + \alpha \sum_{i=1}^N X^T PP^T XWL_i + \beta GW\right)_{jr}}, \quad (9)$$

$$v_{jr} \leftarrow v_{jr} \frac{\left(2CX^T PXW + 2\alpha KW^T\right)_{jr}}{\left(2CVW^T KW + \alpha(A^T + B^T) + 2\beta GV\right)_{jr}}, \quad (10)$$

where $C = H_e^T MH_e$, $A$ represents a $R \times N$ matrix whose rows are $a^T = (diag(X^T PP^T X))^T$ and $B$ denotes a $R \times N$ matrix whose columns are $b = diag(W^T X^T PP^T XW)$.

**2) Fix *others*, update *P* for error correction.**
We can obtain the projection $P$ from the following problem:

$$\min_P J(P) = tr((P^TX - XWV^T)H_e^T MH_e(X^TP - VW^TX^T))$$
$$+ \alpha tr\left(\sum_{i=1}^N (P^T x_i e^T - P^T XW) L_i (P^T x_i e^T - P^T XW)^T\right) \quad .(11)$$
$$+ \beta tr(P^T XGX^T P) + \gamma tr(P^T SP)$$

By taking the derivative $\partial J(P)/\partial P$ of $J(P)$ w.r.t. $P$, setting it to zero, we can update the projection $P$ as

$$P = (X(C + \alpha \Xi + \beta G)X^T + \gamma S)^{-1} XCVW^T X^T, \quad (12)$$

where $\Xi = (E - W)L(E - W)^T$ and $E$ is an $N \times R$ matrix of all ones. After $P$ updated, we can use it together with the factors $W$ and $V$ to compute the adaptive weight matrix $Q$.

**3) Fix *others*, update the adaptive weighting matrix *Q*.**
By removing irrelevant terms to $Q$ from Eq.(7), we can obtain the following reduced formulation:

$$\min_Q J(Q) = \beta tr(H(I-Q)(I-Q)^T H^T), s.t.\ Q \geq 0,\ Q_{ii} = 0, \quad (13)$$

where $H = (X^T P, W, V)^T$. Let $\tau_{ij}$ be the Lagrange multiplier for the nonnegative constraint $Q \geq 0$, and $\Gamma = [\tau_{ij}]$, the Lagrange function $\mathcal{L}_2$ of Eq.(7) can be constructed as

$$\mathcal{L}_2 = \beta tr(H(I-Q)(I-Q)^T H^T) + tr(\Gamma Q^T). \quad (14)$$

By taking the derivative of $\mathcal{L}_2$ w.r.t $Q$, and using the KKT condition $\tau_{ij} q_{ij} = 0$, we can obtain the updating rule for $Q$:

$$q_{ij} \leftarrow q_{ij} \frac{(H^T H)_{ij}}{(H^T HQ)_{ij}}. \quad (15)$$

We summarize the procedures of RFA-LCF in Algorithm 1, where the diagonal matrices $M$ and $S$ are initialized to be the identity matrices as [25] to ensure that each vector $P^i \neq 0$ and $(H_e X^T P - H_e VW^T X^T)^i \neq 0$ over each index $i$ is satisfied.

---

**Algorithm 1:** Our Proposed RFA-LCF Framework
**Inputs:** Training data matrix $X$, control parameters $\alpha, \beta, \gamma$, and the constant $R$ (rank of the factorization).
**Initialization:** Initialize the weight matrix $Q$ using the cosine similarity, i.e., $Q_{ij} = \cos(x_i, x_j)$; Initialize the variables $W$ and $V$ to be random matrices; $t = 0$.
*While not converged do*
1. Update $W$ and representation $V$ by Eqs.(9) and (10);
2. Update the adaptive weight matrix $Q$ by Eq. (15);
3. Update the robust projection $P$ by Eq. (12);
4. Convergence check: if $\|V^{t+1} - V^t\|_F^2 \leq \varepsilon$, stop; else $t = t+1$.
*End while*
**Output:** Learnt new representation $(V^T)^* = (V^T)^{t+1}$.

---

## 4. SIMULATION RESULTS

We conduct simulations on six public real image databases to examine RFA-LCF for data clustering and representation. The results of our RFA-LCF are compared with those of 12 related nonnegative factorization algorithms, i.e., NMF [8], PNMF [28], GNMF [2], DNMF [18], DSNMF [13], PAMGNMF [19], CF [24], LCCF [3], LCF [11], LGCF [9], GRLCF [27] and GCF [26], which are closely related to our RFA-LCF. The information of evaluated datasets are shown in Table I.

Table I. List of used datasets and dataset information.

| Data Type | Dataset Name | # Points | # Dim | # Class |
|---|---|---|---|---|
| Face images | ORL [12] | 400 | 1024 | 40 |
| | UMIST [23] | 1012 | 1024 | 20 |
| | CMU PIE [20] | 11554 | 1024 | 68 |
| Object images | ETH80 [29] | 3280 | 1024 | 80 |
| | COIL100 [14] | 7200 | 1024 | 100 |
| Handwritten images | HWDB1.1-D [30] | 2381 | 196 | 10 |
| | HWDB1.1-L [31] | 12456 | 256 | 52 |

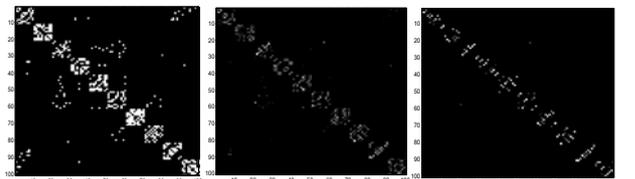

*Fig.1: Visualization comparison of the constructed weights, where (a) Cosine weights, (b) CLR weights and (c) our adaptive weights.*

### 4.1. Visualization of Graph Adjacency Matrix

We compare the adaptive weighting matrix $Q$ of RFA-LCF with the Cosine similarity weights and CLR weights [15] used in

GRLCF. ORL database is used as an example. We choose images of 10 people to construct the adjacency graphs, and the number of nearest neighbors is set to 7 [22] for each method for fair comparison. The weight matrices are shown in Fig.1. We find that more wrong inter-connections are produced in the Cosine weights and CLR weights, which may potentially result in high clustering error, compared with our weights.

### 4.2. Convergence Analysis Results

We present the convergence analysis results of our RFA-LCF in Fig.2. We find that divergence between two consecutive new representations by RFA-LCF is non-increasing in the iteration, and the convergence speed is also fast.

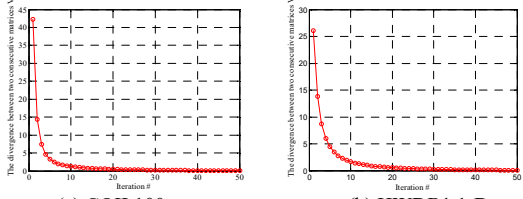

(a) COIL100    (b) HWDB1.1-D
Fig.2: Convergence curves of our RFA-LCF algorithm.

### 4.3. Quantitative Evaluation of Image Clustering

We perform K-means clustering with cosine distance over the new representation of each algorithm. We clearly follow the procedures in [21] for clustering. For each algorithm, the rank R is set to K+1 as [18] and we average results over 30 random initializations for K-means. Accuracy (AC) and F-measure [7] are used for quantitative evaluations. For each database, we vary K from 2 to 10 with step 1, and average the results over 10 random selections of K categories to avoid the bias. Note that the mean and highest AC values are shown in Table II, from which we can find that our RFA-LCF delivers higher AC values than other compared methods in most cases.

### 4.4. Clustering Image Data against Corruptions

We also prepare evaluate the performance of clustering noisy data. To corrupt the data, we add random Gaussian noise with variance being 0-100 with interval 10 into the gray values of selected pixels. The results are shown in Fig.3. The results are obtained on two categories and F-measure is averaged over 50 random selections of categories and $k$-means clustering. We find that our RFA-LCF can outperform the other methods.

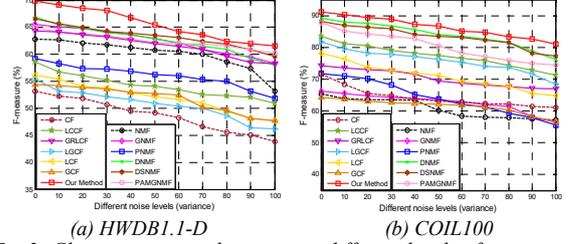

(a) HWDB1.1-D    (b) COIL100
Fig.3: Clustering image data against different levels of corruptions.

## 5. CONCLUSION

We proposed an effective robust flexible auto-weighted local-coordinate concept factorization framework for unsupervised data representation and clustering. Our method improves the accuracy of encoding neighborhood and factorization against noise and outliers by seamlessly integrates the robust flexible CF, robust sparse local coordinate coding and the adaptive weighting. The adaptive weighting strategy avoids the tricky process of selecting optimal parameters in defining the affinity. The flexible residue, coordinate coding and weighting are also performed in the recovered clean data space for potentially enhancing the representation results for clustering. Extensive evaluation have verified the validity of our method. In future, the out-of-sample extension of our model can be investigated.

**Acknowledgement.** This present work is partially supported by National Natural Science Foundation of China (61672365).

Table II. Mean and highest clustering accuracy (AC) over the used six public image databases.

| Dataset / Method | CMU PIE face database | | UMIST face database | | COIL100 object database | |
|---|---|---|---|---|---|---|
| | Mean±std (%) | Best (%) | Mean±std (%) | Best (%) | Mean±std (%) | Best (%) |
| NMF | 23.99±12.51 | 51.2 | 24.36±12.49 | 51.4 | 27.50±11.76 | 52.9 |
| GNMF | 28.10±10.96 | 52.6 | 30.50±12.39 | 58.5 | 31.84±12.09 | 59.6 |
| PNMF | 39.20±11.73 | 60.8 | 46.72±12.34 | 71.6 | 58.14±9.25 | 76.0 |
| DNMF | 31.82±9.01 | 53.5 | 30.86±14.08 | 59.3 | 63.63±13.86 | 90.9 |
| DSNMF | 31.98±11.08 | 57.5 | 38.83±12.78 | 62.5 | 61.85±16.29 | 91.1 |
| PAMGNMF | 36.81±15.53 | 63.3 | 46.04±10.86 | 64.7 | 64.71±13.14 | 89.9 |
| CF | 21.97±13.38 | 51.3 | 21.97±13.38 | 51.3 | 44.61±14.44 | 75.5 |
| LCCF | 37.80±11.25 | 61.9 | 39.01±11.96 | 62.6 | 52.78±16.58 | 85.8 |
| GRLCF | 34.94±13.65 | 62.3 | 43.29±12.14 | 66.5 | 48.03±15.33 | 79.8 |
| LGCF | 36.06±14.07 | 62.5 | 42.61±12.78 | 63.3 | 48.05±15.48 | 80.2 |
| LCF | 37.09±12.41 | 57.3 | 42.15±12.66 | 63.3 | 47.61±15.27 | 79.1 |
| GCF | 35.82±12.14 | 53.3 | 41.57±11.42 | 61.1 | 42.71±12.31 | 67.3 |
| **RFA-LCF** | **41.58±12.84** | **65.9** | **48.83±13.10** | **74.7** | **67.59±13.88** | **92.5** |
| | ETH80 object database | | HWDB1.1-D handwriting | | HWDB1.1-L handwriting | |
| | Mean±std (%) | Best (%) | Mean±std (%) | Best (%) | Mean±std (%) | Best (%) |
| NMF | 24.94±12.42 | 52.3 | 25.55±12.34 | 52.7 | 24.67±11.84 | 51.3 |
| GNMF | 26.42±11.60 | 52.4 | 30.64±11.18 | 54.0 | 27.74±11.40 | 53.6 |
| PNMF | 45.65±13.24 | 72.1 | 35.08±12.73 | 63.4 | 37.51±11.52 | 59.2 |
| DNMF | 22.06±13.33 | 51.2 | 27.02±11.49 | 51.7 | 25.16±11.59 | 51.3 |
| DSNMF | 26.06±12.27 | 54.2 | 33.95±10.10 | 54.7 | 27.93±12.15 | 54.3 |
| PAMGNMF | 25.49±12.41 | 52.5 | 28.73±11.63 | 53.1 | 25.39±12.33 | 52.4 |
| CF | 33.65±13.23 | 63.5 | 28.99±12.01 | 55.3 | 28.66±11.21 | 53.7 |
| LCCF | 34.97±14.58 | 67.9 | 32.74±12.58 | 58.5 | 30.62±11.47 | 54.1 |
| GRLCF | 36.39±13.61 | 66.8 | 34.07±12.59 | 60.8 | 32.79±11.72 | 57.1 |
| LGCF | 36.34±13.42 | 66.0 | 34.95±12.88 | 62.1 | 32.63±11.83 | 56.6 |
| LCF | 36.04±13.79 | 67.1 | 29.17±11.91 | 54.9 | 31.43±11.74 | 56.1 |
| GCF | 33.40±13.29 | 64.0 | 28.89±11.97 | 55.0 | 30.52±11.16 | 54.2 |
| **RFA-LCF** | **49.05±14.63** | **78.0** | **38.71±13.59** | **68.7** | **39.85±11.83** | **62.2** |